\documentclass[sigconf]{acmart}

\AtBeginDocument{%
  \providecommand\BibTeX{{%
    \normalfont B\kern-0.5em{\scshape i\kern-0.25em b}\kern-0.8em\TeX}}}

\renewcommand\footnotetextcopyrightpermission[1]{} 
\usepackage[caption=false,font=normalsize,labelfont=sf,textfont=sf]{subfig}
\usepackage{bbding}
\usepackage{multirow} 
\usepackage{caption}
\usepackage{footnote}
\usepackage{shadowtext}
\usepackage{colortbl}
\usepackage{graphicx}
\usepackage{stfloats}
\usepackage[capitalize]{cleveref}

\begin{document}

\title{An Improved Graph Pooling Network for Skeleton-Based \\Action Recognition}

\author{Cong Wu}
\affiliation{%
  \institution{Jiangnan University}
  \city{Wuxi}
  \country{China}}
\email{congwu@stu.jiangnan.edu.cn}

\author{Xiao-Jun Wu}
\affiliation{%
  \institution{Jiangnan University}
  \city{Wuxi}
  \country{China}}
\email{wu\_xiaojun@jiangnan.edu.cn}

\author{Tianyang Xu}
\affiliation{%
  \institution{Jiangnan University}
  \city{Wuxi}
  \country{China}}
\email{tianyang\_xu@163.com}

\author{Josef Kittler}
\affiliation{%
  \institution{University of Surrey}
  \city{Guildford}
  \country{United Kingdom}}
\email{j.kittler@surrey.ac.uk}

\begin{abstract}
Pooling is a crucial operation in computer vision, yet the unique structure of skeletons hinders the application of existing pooling strategies to skeleton graph modelling. In this paper, we propose an Improved Graph Pooling Network, referred to as IGPN. The main innovations include: Our method incorporates a region-awareness pooling strategy based on structural partitioning. The correlation matrix of the original feature is used to adaptively adjust the weight of information in different regions of the newly generated features, resulting in more flexible and effective processing. To prevent the irreversible loss of discriminative information, we propose a cross fusion module and an information supplement module to provide block-level and input-level information respectively. As a plug-and-play structure, the proposed operation can be seamlessly combined with existing GCN-based models. We conducted extensive evaluations on several challenging benchmarks, and the experimental results indicate the effectiveness of our proposed solutions. For example, in the cross-subject evaluation of the NTU-RGB+D 60 dataset, IGPN achieves a significant improvement in accuracy compared to the baseline while reducing Flops by nearly 70\%; a heavier version has also been introduced to further boost accuracy.
\end{abstract}



\maketitle

\begin{table*}[t]
\caption{Performance comparison with different backbones on cross-subject of NTU-RGB+D 60.
The inference speed and GPU memory is measured on a single GeForce RTX 2080 Ti.
}
\centering
\scalebox{0.92}{
\begin{tabular}{c c c c c c c}
\hline
{Baseline} &{Pooling} & {Flops (G)} &{Infer Speed (Videos per Second)} &{GPU Mem (MiB)} & {Acc (\%)} \\
\hline
\multirow{3}{*}{AGCN~\cite{shi2019two}} &-  &4.17 &589 &6988 &88.9\\
                    &IGP-Light &1.16  &915 &4031 &90.1 \\
                      & IGP-Heavy  &5.02  &515 &7195 &90.5 \\ \hline
\multirow{3}{*}{Graph2Net~\cite{wu2021graph2net}} &-  &1.16&824 &9273 &89.1 \\
                    & IGP-Light &0.44  &1268 &4689 &90.2 \\
                      & IGP-Heavy &1.28  &687 &8463 &91.2\\ \hline
\multirow{3}{*}{CTR-GCN~\cite{chen2021channel}} &- &1.98 &75 &9613 &89.9 \\
                        & IGP-Light &0.70 &115 &4573 &90.9  \\
                      &  IGP-Heavy &2.04 &61 &8229 &91.1 \\ 
\hline
\end{tabular}}
\label{table_}
\end{table*}

\section{Introduction}
Recent studies have verified the inherent advantages of Graph Convolutional Networks (GCNs) in the representation of non-Euclidean structured data~\cite{wu2020comprehensive}, with GCN-based methods achieving leading performance in various fields~\cite{gao2019graphtracking,liu2022deep,huang2022dual}.
Yan \emph{et al.}~\cite{yan2018spatial} first proposed to construct spatial-temporal graph to model skeleton sequences according to their natural arrangement. Refer to this paradigm, subsequent methods~\cite{shi2019two, si2019attention, liu2020disentangling, yussif2023self, chen2021channel, lee2023hierarchically,chen2023occluded} have made significant improvements in performance.
However, only global pooling is used before the final classifier in most methods, leading to the redundancy of homogeneous information in graph modelling~\cite{qi2022semantic}.
Simultaneously, pooling is a conventional feature aggregation technique in Convolutional Neural Networks (CNNs), which typically integrates local information obtained after convolutional layers to produce more abstract feature representations. 
This operation reduces redundant information, lowers the computational cost, prevents over-fitting, and enhances the model's generalisation ability, while also increasing the integration of features and obtaining higher-level abstract representations through aggregation.
The pooling operation is not only justified theoretically~\cite{boureau2010theoretical,bianchi2024expressive}, 
but also has been shown to significantly improve performance in many computer vision tasks~\cite{scherer2010evaluation,hou2020strip}.

Skeleton graph pooling strategy has recently gained significant attention in research.
\cite{aberman2020skeleton} introduced novel pooling and unpooling operators within different homeomorphic skeletons. 
\cite{chen2020graph} divided the skeleton into regions based on physical arrangement and performed continuous pooling operations in each local region.
Those approaches bring higher computational efficiency while achieving higher recognition accuracy.
However, both methods rely on fixed manual pooling rules, which may remove useful information and limit their flexibility.
\cite{peng2021tripool} introduced triplet pooling loss to facilitate adaptive clustering of skeleton graph nodes and automatically generate new embedded matrix representations. This approach introduces great flexibility, making the pooling process a trainable method. However, it does not explicitly consider the structural-semantic information.
In addition, existing pooling methods tend to force the network to discard essential but not the most critical nodes gradually.
However, the accumulation of multiple pooling operations can significantly affect the final performance. Hence, it is necessary to preserve and strengthen the discriminative and heterogeneous features while discarding unnecessary homogeneous features.

To address the limitations of existing pooling networks that rely on either fixed manual rules or excessive flexibility while ignoring the inherent structure, we propose a trainable structure pooling operation that strikes a balance between flexibility and structural preservability.
Specifically, we define the pooling area according to the physical structure of the skeleton, then calculate the contribution between each feature point in the pooling area and the entire graph representation according to the embedding similarity. After that, we adjust the proportion of the current feature point in the pooled features by constructing a trainable assign matrix. 

With the accumulation of multi-layer pooling operations, the loss of information will only intensify without remediation, the network capacity also will be surplus.
Here we propose block-based and network-based feature enhancement strategies, namely cross fusion block and information supplement module.
In cross fusion block, based on the original pooling network, we design another parallel operation that directly performs graph operations and then fuses the features obtained from the two parallel structures after alignment. 
In this way, feature fusion of different granularities is performed to supplement and strengthen the information of features.
In information supplement module, we decompose the skeleton sequence into position and vector-based features, then project them into common space through graph embedding for sufficient alignment, and combine them to obtain an efficient representation through feature fusion.

The conducted ablation experiments demonstrate our motivation and the effectiveness of the proposed method, and the results combined with various mainstream methods consistently show that our method can significantly reduce computational overhead while improving model performance. 
As shown in Table~\ref{table_}, 
the proposed improved graph pooling operation 
can can greatly reduce training overhead while ensuring improvement in accuracy performance.

Overall, our innovations can be summarised as follows:
\begin{itemize}
\item 
Aiming at the structural properties of the skeleton sequences, we propose a structure pooling strategy with region awareness, which increases the pooling flexibility while maintaining the skeleton structural characteristics.
\item 
We use the cross fusion module and information supplement module to provide block-level and input-level information respectively, thus avoiding the loss of a large amount of information and the under-saturation of the network.
\item We conduct comprehensive experiments on NTU-RGB+D (60\&120)~\cite{shahroudy2016ntu,liu2019ntu}, UWA3D Multiview Activity II~\cite{rahmani2016histogram} datasets respectively, the  experimental findings provide full validation of our strategies.
\end{itemize}

\section{Related Work}
\subsection{Skeleton-based Action Recognition}
\noindent\textbf{Non-GCN Methods.} 
Due to the particularity of the structure of skeleton features, existing strategies often focus on their essential physical characteristics to carry out targeted modelling. 
Traditional methods for skeleton sequence processing and recognition generally involve deforming the skeleton sequence by constructing suitable descriptors.
\cite{chen2006human} used the star skeleton as a representative descriptor of the human skeleton, then converted the skeleton sequence into a symbol sequence that can be processed by the Hidden Markov Model. \cite{vemulapalli2014human} proposed to represent skeleton sequences as curves in the Lie group, and the classification was performed by combining classifiers. 
\cite{han2010discriminative} proposed using Hierarchical Gaussian Process Latent Variable Model to learn a hierarchical manifold space of motion patterns. 
With the development of research in recent years, the effectiveness of deep learning frameworks has been widely verified in various vision tasks~\cite{hochreiter1997long,mikolov2010recurrent,he2016deep}, which inspired people to leverage the discriminative feature extraction capabilities of those deep networks for skeleton modelling.
Du \emph{et al.} \cite{du2015skeleton} proposed a method that converts the skeleton sequence into an image format, which can be processed using CNNs for feature extraction. 
\cite{liu2017enhanced} further used colourisation to implicitly encoding the spatiotemporal information of the skeleton sequences. 
\cite{ke2017new} generated multiple sets of clip-based representations and processed this information in parallel with the help of the Multi-Task Learning Network.
Inspired by the natural advantages of Recurrent Neural Networks (RNNs) and Long Short-Term Memory (LSTM) in sequence modelling, methods such as~\cite{du2015hierarchical, song2017end, avola20192, li2021memory} used RNNs or LSTMs to capture long-term contextual information.

\noindent\textbf{GCN-based Methods.}
As a typical message-passing network, the emergence of GCNs provides a new paradigm for skeleton sequence modelling.
Yan \emph{et al.}~\cite{yan2018spatial} extended the basic GCNs operations to the spatiotemporal dimension. Shi \emph{et al.}~\cite{shi2019two} moved away from the fixed graph modelling paradigm and proposed an adaptive relational representation. 
\cite{liu2020disentangling} proposed a complex disentangling and unifying operation based on the previous works, which significantly improved the model's classification performance. \cite{chen2021channel} constructed a more comprehensive graph learning framework by strengthening the representation granularity of channels.
Although the skeleton sequence contains relatively compact information and requires low computational overhead, researchers have recently conducted extensive research on recognition efficiency to avoid blindly stacking models.
Combined with high-level semantic information, Zhang \emph{et al.}~\cite{zhang2020semantics} proposed a simple yet effective semantics-guided neural network. 
\cite{cheng2020skeleton} used efficient shift operations to implement the interaction of spatiotemporal information, significantly reducing the number of trainable parameters and computational complexity. 
Taking into account both accuracy and efficiency, Wu \emph{et al.}~\cite{wu2021graph2net} divided the channel dimensions and endowed them with diverse modelling capabilities, thereby improving the utilisation of network parameters.
Qin~\emph{et al.}~\cite{qin2022fusing} proposed a novel approach for constructing high-dimensional feature representations using angular encoding.
The study by Liu \emph{et al.}~\cite{liu2020adversarial} proposed a successful attack on existing skeletal graph models by exploiting spatiotemporal constraints.

\subsection{Graph Pooling Operation}
In contrast to digital images, skeletons are typical non-Euclidean structural data with feature points that are not rigidly arranged and exhibit strong physical structure characteristics. As a result, pooling methods designed for image tasks, which typically involve selecting a regular rectangular area and obtaining a new feature point by averaging or taking the maximum value of all points in that area, are not directly applicable.
Instead, skeleton graph pooling often needs to consider both the graph's structure and the arrangement of its nodes. Specifically, it includes how to select the pooling area and how to generate a new feature representation.
There are three main approaches for graph pooling: hard rule, graph coarsening, and node selection.
Hard rule involves pre-setting pooling nodes and rules based on the known graph structure. 
Chen \emph{et al.}~\cite{chen2020graph} proposed a structure-based graph pooling scheme that considers the unique structure and prior topological knowledge of skeleton sequences.
Graph coarsening is a trainable version of the hard rule that involves clustering nodes and synthesising a super-node.
A differentiable graph pooling module constructed by \cite{ying2018hierarchical} to map the nodes of each layer to a set of clusters. 
\cite{bianchi2020spectral} introduced a graph-based clustering method for graph pooling using continuous relaxation of the normalised minimum segmentation problem and achieved cluster assignment by training Graph Neural Networks (GNNs). \cite{peng2021tripool} proposed a hierarchical graph representation method using triplet pooling loss.
Node selection is to select some critical nodes to construct a new graph representation, thereby replacing the original structure. \cite{gao2019graph} proposed graph pooling and unpooling operations by selecting important nodes and restoring global nodes. \cite{lee2019self} used the attention mechanism in GCNs to generate graph topology masks and used this mask to complete the pooling operation. 
In most previous studies, people often directly apply the graph pooling strategy to the skeleton task, failing to consider the structural properties of skeleton features and the specificity of different features.

\section{The Proposed Approach}
In this section, we will begin by constructing the basic graph modelling module. Subsequently, the proposed innovations will be introduced step by step. Finally, we will describe the overall architecture that integrates all the components.

\subsection{Basic Graph Modelling Module}
\label{section3.1}
Referring to the standard graph definition, for any given graph $G=(V, E)$, $V$ represents the set of nodes, $E$ represents the set of corresponding edges.
During the modelling process, each node aggregates information from its neighbouring nodes and updates its features accordingly.
Given the graph $G_{l}$ at $l$\_th layer, according to the defined rules, the 
newly generated graph can be expressed as,
\begin{equation}
G_{l+1}=\operatorname{F}(G_{l},W_{l}) 
	   =\operatorname{F^{U}}((\operatorname{F{^{A}}}(G_{l},W_{l}^{A})),W_{l}^{U}),
\label{eqno1}
\end{equation}
$\operatorname{F}$ represents the graph operation, $W$ denotes the weight. The graph operation can be divided into two main components, namely aggregation $\operatorname{F^{A}}$ and updating $\operatorname{F^{U}}$, where $W^{A}$ and $W^{U}$ represent the corresponding weights.

As described in \cite{yan2018spatial}, spatial graph convolution involves performing a standard 2D convolution with a $1\times1$ kernel, followed by multiplying the resulting feature with a normalised adjacency matrix. 
Meanwhile, temporal graph convolution can be represented as a well-defined convolution operation on the constructed temporal graphs. 
In this work, we adopt existing methods as the backbone network to fairly measure the effectiveness of the proposed method.

\subsection{Structure Pooling Strategy with Region Awareness}
\label{section3.2}

In this section, we propose a novel structure pooling strategy (SSP) that incorporates region awareness to achieve more flexibility while preserving the underlying structure of the skeleton graph.
The graph convolution operation applied to skeleton-based action recognition typically involves both the feature itself and its corresponding relationship representation (usually represented by its adjacency relationship). 
Therefore, to perform pooling on the skeleton sequence, it is essential to not only select the pooling area but also take into account the transformation between the old and new graphs.
Additionally, it is necessary to generate a new relationship descriptor for the features that have undergone the pooling operation.

\begin{figure}[t]
\centering
\includegraphics[width=1.\linewidth]{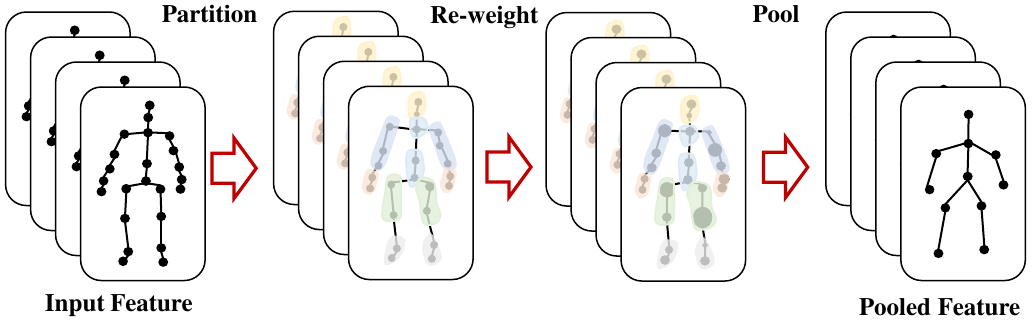}
\caption{{Structural Spatial Pooling.} 
We divide the skeleton according to its spatial structure and assign different attentions to different nodes to achieve adaptive structured spatial pooling.
The size of the dot 
indicates the corresponding intensity of attention.
}
\label{fig1}
\end{figure}
\begin{figure}[t]
\centering
\includegraphics[width=1.\linewidth]{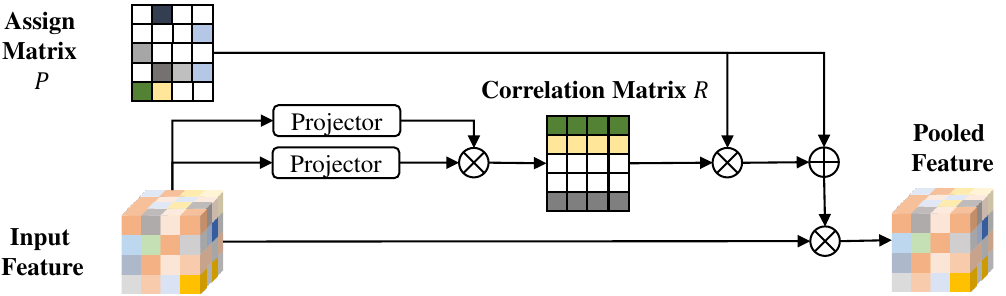}
\caption{{Structure pooling strategy with region awareness.} We combine the characteristics of the current feature itself, design an adaptive feature pooling operation, and automatically calculate the relationship matrix of the current feature.}
\label{fig2}
\end{figure}
We begin by denoting the current features at $t$\_th frame as $x_t=[x_{t0}, x_{t1},...,x_{ti},...,x_{tN}]$, where $x_{ti}\in \mathbf{R}^C$ corresponds to the feature of node $v_{i}$ at frame $t$. We then divide the skeleton graph, which consists of $N$ nodes, into $M$ regions based on its physical structure, and then perform feature aggregation separately on each region to obtain a new graph with $M$ nodes.
We generate the fundamental transformation matrix $P\in \mathbb{R}^{N\times M}$ between the old and new features. If joint $i$ belongs to the $j$\_th partitioned area (i.e., corresponds to the feature point $j$ in the new graph), then $P_{ij}=1$; otherwise, $P_{ij}=0$.
While we believe that the above assignment setting is conducive in preserving structural characteristics, it ignores the specificity responses of each point across the entire graph or even across different input sequences.
For example, 'Rub Two Hands' is more focused on hand movement, while 'Jump Up' requires coordination of the entire body.

To tackle these challenges, we have imbued the pooling operation with the requisite flexibility without compromising its structural integrity. 
The entire calculation process is illustrated in Figure~\ref{fig1} and Figure~\ref{fig2}. 
Initially, we compute the correlation matrix $R$, which captures the correlation between each feature point and other locations.
The value associated with each feature point can be represented as the average of its connections with other feature points in the current graph, that is,
\begin{equation}
\forall i\in N, R_{ti}=\sigma (\sum_{j=1}^{N} \frac{\phi({h(x_{ti})})^{T}\psi({h(x_{tj})})}{N}),
\label{eqno3}
\end{equation}
where $\sigma(\cdot)$ is the normalisation operation, and $\phi$ and $\psi$ are projection functions that can be implemented by convolution filters, 
the dimension of project space is set as $C/r$. 
The entry $R_{ti}$ measures the mean of the connection between point $v_{i}$ and all other nodes at $t$\_th frame, 
and this can be computed based on their corresponding latent representations $h(\cdot)$.
After that, the new feature after the spatial pooling operation can be expressed as,
\begin{equation}
\begin{split}
\operatorname{SPooling}(x_t)&=x_tR_tP,
\label{eqno4}
\end{split}
\end{equation}
Residual connection is also introduced to achieve training stability.
That is, for each newly generated $x^{sp}_{tj}$,
\begin{equation}
x^{sp}_{tj}=\sum_{i=1}^{N}x_{ti} (P_{ij} + R_{ti} P_{ij}).
\label{eqno5}
\end{equation}
Combing with (\ref{eqno3}), we have
\begin{equation}
x^{sp}_{tj} = \sum_{i=1}^{N}x_{ti} (P_{ij} + \sigma(\sum_{j=1}^{N} \frac{\phi({h(x_{ti})})^{T} \psi({h(x_{tj})})}{N}) P_{ij}).
\label{eqno6}
\end{equation}
By introducing a trainable allocation matrix, our proposed strategies provide greater flexibility in treating each feature differently during the pooling operation. 
Furthermore, we retain the basic transformation form, which ensures the preservation of the structural characteristics of the skeleton sequence.
Finally, to obtain the relation representation for the newly pooled feature, we need to compute a new adjacency matrix that reflects the physical structure of the new graph. Since we maintain the basic structure in this pooling process, that is, the pooling is performed in a specific area, the feature representation and its physical structure associations can be conveniently obtained based on the pooling strategy. This enables the generation of the corresponding relationship description for each newly generated feature.

\begin{figure*}[t]
\centering
\includegraphics[height=0.32\textheight]{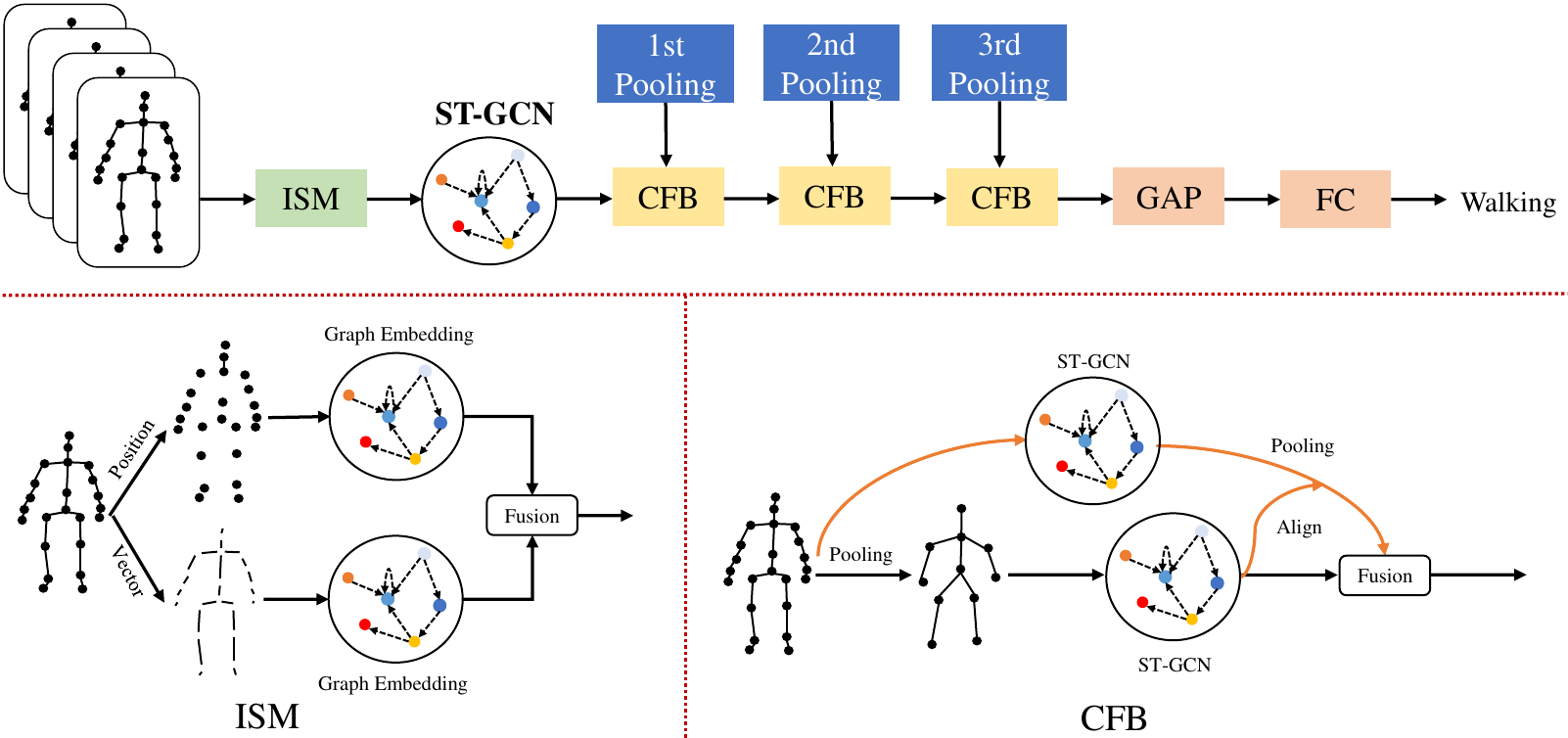}
\caption{{The Overall Structure.} 
ISM stands for Information Supplement Module, CFB stands for Cross Fusion Block, and GCN, GAP, and FC correspond to graph convolution operations, global average pooling, and fully connected layers, respectively.
}
\label{fig3}
\end{figure*}

As for temporal pooling, due to the regular arrangement of the skeleton on the temporal dimension as discussed in Section~\ref{section3.1}, we can directly apply conventional pooling strategies.
The final pooled feature is obtained by performing an average pooling with a kernel of 2 on the current feature, as shown in  Figure~\ref{fig2}, so the final formal can be written as,
\begin{equation}
\begin{split}
x^{p}&=\operatorname{STPooling}(x)=\operatorname{TPooling}(\operatorname{SPooling}(x))  \\
&=\operatorname{TAP}(x^{sp})=(x^{sp}_{[0::2]}+x^{sp}_{[1::2]})/2.
\end{split}
\label{eqno7}
\end{equation}
$\operatorname{TAP}$ means the Average Pooling on the temporal dimension with a kernel size of 2, $[0 :: 2]$ and $[1 :: 2]$ represent to extract 1 frame every 2 frames starting from frame 0 and 1 respectively.

\subsection{Cross Fusion Block}
\label{section3.3}
The pooling operation significantly enhances the gathering and transmission of features and has the potential to reduce redundancy.
Bu tit may also have the opposite effect for skeletons due to
the irreparable and significant loss of effective information.
To alleviate the information loss during the pooling process, we maintain the basic pooling structure and construct a cross fusion block in a plug-and-play form.
As depicted in  Figure~\ref{fig3}, the lower branch of the Cross Fusion Block represents the original graph pooling network module, while the newly introduced upper branch models obtain finer skeleton features that are cross-fused with the lower branch to enhance the representation ability of the original features.

As a fundamental component, we construct a graph pooling network module following the definitions in Section~\ref{section3.1} and Section~\ref{section3.2}; Specifically, we first obtain a new graph using the predefined pooling strategy and then perform a graph convolution operation on the newly generated feature representation. 
Specially,
for the current input feature of this block, based on the feature $x_{p}$ obtained by pooling in Equation~\ref{eqno7}, the new representation can be expressed as,
\begin{equation}
h=\operatorname{GraphConv}(x^{p}), \\
\label{eqno8}
\end{equation}
where $\operatorname{GraphConv}$ general represents the graph convolution operation.
However, with this process, multiple nodes in the pooling area are replaced by a single new node representation before the graph operation. 
As a consequence, the subsequent graph convolution is executed on the compressed skeleton features with much lower dimension, resulting in the modelling of more abstract features. However, finer parts may not be fully captured in this process. Therefore, we need to emphasise the modelling of more refined features for our pooling network.
We first use graph convolution to further model the current features $x$ to obtain a more informative feature representation $\hat{x}$,
\begin{equation}
\hat{x}=\operatorname{GraphConv}(x).
\label{eqno10}
\end{equation}
Then, the fusion of $\hat{x}$ with $h$ is performed to improve the overall representation ability. However, as $\hat{x}$ and $h$ have different resolutions, there is a misalignment issue between them.
To address this, we align the dimensions of these two features by performing a pooling operation on $\hat{x}$ using $h$ as a reference. Finally, the two aligned features are fused, as illustrated in  Figure~\ref{fig3}.
According to the previous definition, we denote the upper branch feature after alignment as $e$,
\begin{equation}
\begin{split}
e=\operatorname{STPooling}(\hat{x})=\operatorname{TPooling}(\operatorname{SPooling}(\hat{x}))
\end{split}
\label{eqn11}
\end{equation}
this formula represents the alignment operation, where $R(\hat{x})$ represents the relational representation calculated based on the current feature $\hat{x}$, and the calculation process can be referred to Equation~\ref{eqno3}.
After that, the specific operations in the entire cross fusion block can be expressed as,
\begin{equation}
y=s\cdot h+(1-s)\cdot e
\label{eqno12}
\end{equation}
where $s$ means the corresponding weight, we will discuss it later in experimental section.

\subsection{Information Supplement Module}
\label{section3.4}
Given that the introduction of the pooling network significantly eliminates most information, it is reasonable to believe that the current network structure has not fully utilised its performance potential.
Rather than using conventional techniques such as distillation or pruning to compress the current network size, we opt to enhance the input features to fully exploit the modelling capabilities of the existing network structure.
Previous related methods often adopted complex multi-stream structures~\cite{shi2019two, liu2020disentangling, cheng2020skeleton} to exploit the multi-dimensional representation information of input features. However, this strategy usually results in limited performance improvement while significantly increasing the computational and parameter requirements, thereby limiting the advantages of using skeleton features in real-world application deployment.

Compared to traditional methods, our approach involves performing appropriate pre-processing on the input side of the network to construct efficient input features while maintaining the feature extraction network, as illustrated in  Figure~\ref{fig3}. We focus on two critical feature descriptors for the skeleton sequence: position-based and vector-based features, corresponding to joints and bones in the skeleton.
To align and integrate the position-based and vector-based features of the skeleton sequence, we use graph convolution to project them into a high-dimensional vector space. The projection operation maps each feature into a common vector space, enabling direct alignment and fusion. This results in a more informative representation that can capture both the joint and bone information.
These processes can be expressed as follows,
\begin{equation}
f_p(x):x\underset{position}{\rightarrow}x^{pos},
\nonumber
\end{equation}
\begin{equation}
f_v(x):x\underset{vector}{\rightarrow}x^{vec}.
\label{eqno13}
\end{equation}
$x^{pos}$ and $x^{vec}$ correspond to the position-based and vector-based features obtained from the initial skeleton sequence, respectively, as mentioned in~\cite{shi2019two}.
\begin{equation}
x^{vec}_{in} = \operatorname{GrpahConv}(\operatorname{Norm}(x^{vec})),
\nonumber
\end{equation}
\begin{equation}
x^{pos}_{in} = \operatorname{GrpahConv}(\operatorname{Norm}(x^{pos})).
\label{eqno14}
\end{equation}
The normalisation operation can be expressed as,
\begin{equation}
\operatorname{Norm}(x)=\frac{x-E(x)}{\sqrt{var(x)+\epsilon}}*\gamma+\beta,
\label{eqno15}
\end{equation}
where $\gamma$ and $\beta$ are trainable parameters, $E$ and $var$ are the mean and standard-deviation.
The final input $x_{in}$ is the fusion of position-based feature and vector-based feature.
\begin{equation}
x_{in} = 
\operatorname{Concat}(x^{vec}_{in},x^{pos}_{in}),
\label{eqno16}
\end{equation}
where $\operatorname{Concat}$ means feature concatenation along the channel dimension. 
The proposed input-strengthening strategy brings only a limited increase in the number of parameters and computational overhead compared to the original network structure but greatly enhances the representation of the new input features. The effectiveness of this strategy is verified through follow-up experiments.

\subsection{The Overall Structure}
Combining the above descriptions, we construct the improved pooling network model starts with a baseline backbone taken from~\cite{chen2020graph}, and replaces the basic graph convolution operation with the module proposed in~\cite{wu2021graph2net}. We also make some optimisations to the data processing part following~\cite{chen2021channel}. Our innovations are then added to the baseline, resulting in the improved pooling network model whose final structure is shown in  Figure~\ref{fig3}. The improved pooling network model begins by processing the input feature using an information supplement module to obtain a new representation with richer information. The output of this module is then passed through several cross fusion blocks, where each block combines a graph pooling and a graph convolution operation. Finally, the classification results are obtained after global pooling and a classification layer.
We also remove the cross fusion block to construct a lightweight version, named \textbf{IGPN-Light}. For distinction, the default model is denoted as \textbf{IGPN-Heavy}.

\section{Experiments}
\subsection{Datasets and Implementation Details} 

\noindent\textbf{NTU-RGB+D.} 
The NTU-RGB+D 60 dataset~\cite{shahroudy2016ntu} comprises 56,880 videos captured from 80 camera views and 40 subjects, with 60 action categories. The NTU-RGB+D 120 dataset~\cite{liu2019ntu} extends the previous version to 120 action categories, with 114,480 videos captured from 155 views and 106 subjects.
To assess the effectiveness of our approach, we utilise the commonly adopted evaluation methods: cross-subject and cross-view for NTU-RGB+D 60,
cross-subject and cross-set for NTU-RGB+D 120.
For cross-subject evaluation, we use videos performed by half of the subjects for training and the remaining half for testing in both datasets. 
For cross-view evaluation, we use the samples collected by cameras with ID 2 and 3 for training and the remaining for testing in NTU-RGB+D 60. 
For cross-set evaluation, we use samples with even camera IDs for training and the others for testing in NTU-RGB+D 120.

\noindent\textbf{UWA3D Multiview Activity II.} 
The UWA3D Multiview Activity II Dataset~\cite{rahmani2016histogram} comprises 30 different classes performed by 10 characters, with each video captured by four different views (front, left, right, and top).
We randomly select two views as training data and test on the remaining two views separately. In our experiments, we report experimental results for all combinations of views, as well as the final average performance.

\begin{table}[!ht]
\centering
\caption{The implementation details of the proposed method.}
\scalebox{0.92}{
\begin{tabular}{c|c c}
\hline 
Dataset &NTU-RGB+D &UWA3D \\
\hline
Optimizer  &\multicolumn{2}{c}{SGD, Momentum=0.9, Nesterov=True} \\
Epoch &65 & 100  \\
Warm up epoch &\multicolumn{2}{c}{5}  \\
Learning rate &\multicolumn{2}{c}{1e-1}   \\
Learning rate policy &\multicolumn{2}{c}{Linear decay with 0.1}\\
Decay step &[35,55]  &[50,80]\\
Weight decay &0.0004  &0.01\\
Batchsize &64  &32\\
Frame &64 &60 \\
Data augmentation & \multicolumn{2}{c}{Random rotate} \\
\hline
\end{tabular}
}
\label{table:detail}
\end{table}

\subsection{Implementation Details}
\noindent\textbf{Training and evaluation.} 
In general, our experiments are performed on the PyTorch platform, with cross-entropy loss and Stochastic Gradient Descent with Nesterov Momentum (0.9) employed as the training loss and optimisation algorithm, respectively. 
We use the processing steps from~\cite{zhang2020semantics} for the NTU-RGB+D datasets. The initial learning rate is set to 0.1, with a batch size of 64 for NTU-RGB+D and 32 for UWA3D Multiview Activity II dataset. 
For NTU-RGB+D datasets, the learning rate is reduced to 1/10 of the previous epoch at the 35th and 55th epoch, and training stopped at the 65th epoch. 
For UWA3D Multiview Activity II dataset, the learning rate decay at 50th and 80th epochs, and the training stopped at the 100th epoch. 
We apply a warm-up strategy for the first five epochs across all datasets. 
Please find the details from Table~\ref{table:detail}.

\begin{figure}[h]
\centering
\begin{tabular}{@{\extracolsep{\fill}}c@{}c@{\extracolsep{\fill}}}
\includegraphics[width=1.\linewidth]{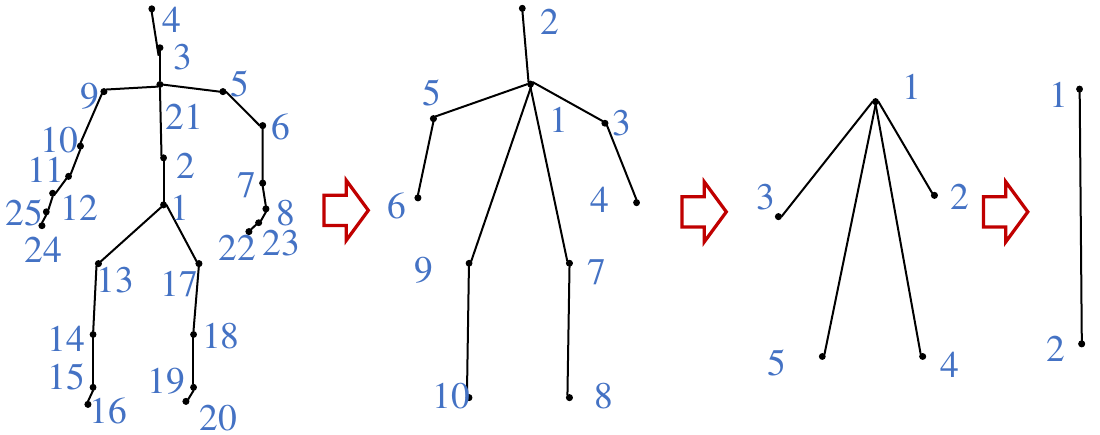} 
\\
(a) For NTU-RGB+D datasets.  \\
\includegraphics[width=1.\linewidth]{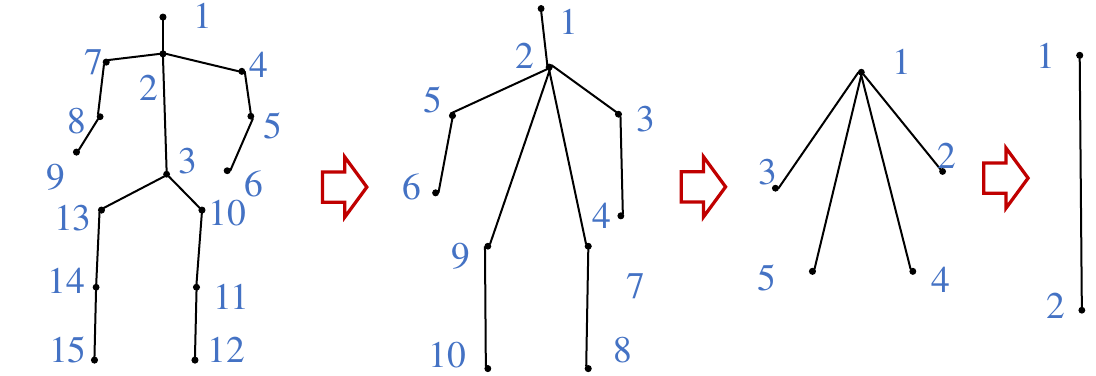}\\
(b) For UWA3D Multiview Activity II dataset.\\
\end{tabular}
\caption{The process of spatial pooling.}
\label{fig:fig1}
\end{figure}

\begin{table}[t]
\caption{\textbf{Predefined spatial pooling rules.} 
}
\centering
\subfloat[For NTU-RGB+D datasets.]{
\scalebox{0.92}{
\begin{tabular}{c c|c c|c c}
\hline
\multicolumn{2}{c|}{\textbf{First Pooling}} & \multicolumn{2}{c|}{\textbf{Second Pooling}} &\multicolumn{2}{c}{\textbf{Third Pooling}}\\ \hline
\textbf{Old IDs} &\textbf{New ID}	& \textbf{Old IDs} &\textbf{New ID} 	& \textbf{Old IDs} &\textbf{New ID}  \\
\hline
(1,2,21) &1 &(1,2)&1&(1,2,3)&1\\
(3,4,21)  &2 &(3,4)&2&(4,5)&2\\
(5,6,7) &3 &(5,6)&3\\
(8,22,23) &4 &(7,8)&4\\
(9,10,11) &5 &(9,10)&5\\
(12,24,25) &6&&\\
(13,14) & 7&&\\
(15,16) &8&& \\
(17,18)&9&& \\
(19,20)&10&&\\
\hline
\end{tabular}}}
\\
\subfloat[For UWA3D Multiview Activity II dataset.]{
\scalebox{0.92}{
\begin{tabular}{c c|c c|c c}
\hline
\multicolumn{2}{c|}{\textbf{First Pooling}} & \multicolumn{2}{c|}{\textbf{Second Pooling}} &\multicolumn{2}{c}{\textbf{Third Pooling}}\\ \hline
\textbf{Old IDs} &\textbf{New ID}	& \textbf{Old IDs} &\textbf{New ID} 	& \textbf{Old IDs} &\textbf{New ID}  \\
\hline
(1,2) &1 &(1,2)&1&(1,2,3)&1\\
(2,3)  &2 &(3,4)&2&(4,5)&2\\
(4,5) &3 &(5,6)&3\\
(5,6) &4 &(7,8)&4\\
(7,8) &5 &(9,10)&5\\
(8,9) &6&&\\
(10,11) & 7&&\\
(11,12) &8&& \\
(13,14)&9&& \\
(14,15)&10&&\\
\hline
\end{tabular}
}
}
\label{table:relue}
\end{table}

\noindent\textbf{Predefined spatial pooling rules.} 
As shown in the Figure~\ref{fig:fig1}, we give the spatial skeleton graph in the whole network. The left side of the arrow is the spatial graph before pooling, and the right side of the arrow represents the newly generated graph after the structure spatial pooling operation.

As for the pooling rules, for the given skeleton representations, structured region division is performed first; with all nodes in each region condensed, a new sparse and compact graph representation can be obtained. 
Refer to~\cite{peng2021tripool}, the flow of several pooling operations is shown in the Table~\ref{table:relue}.

\begin{figure}[t]
\centering
\includegraphics[width=0.475\textwidth]{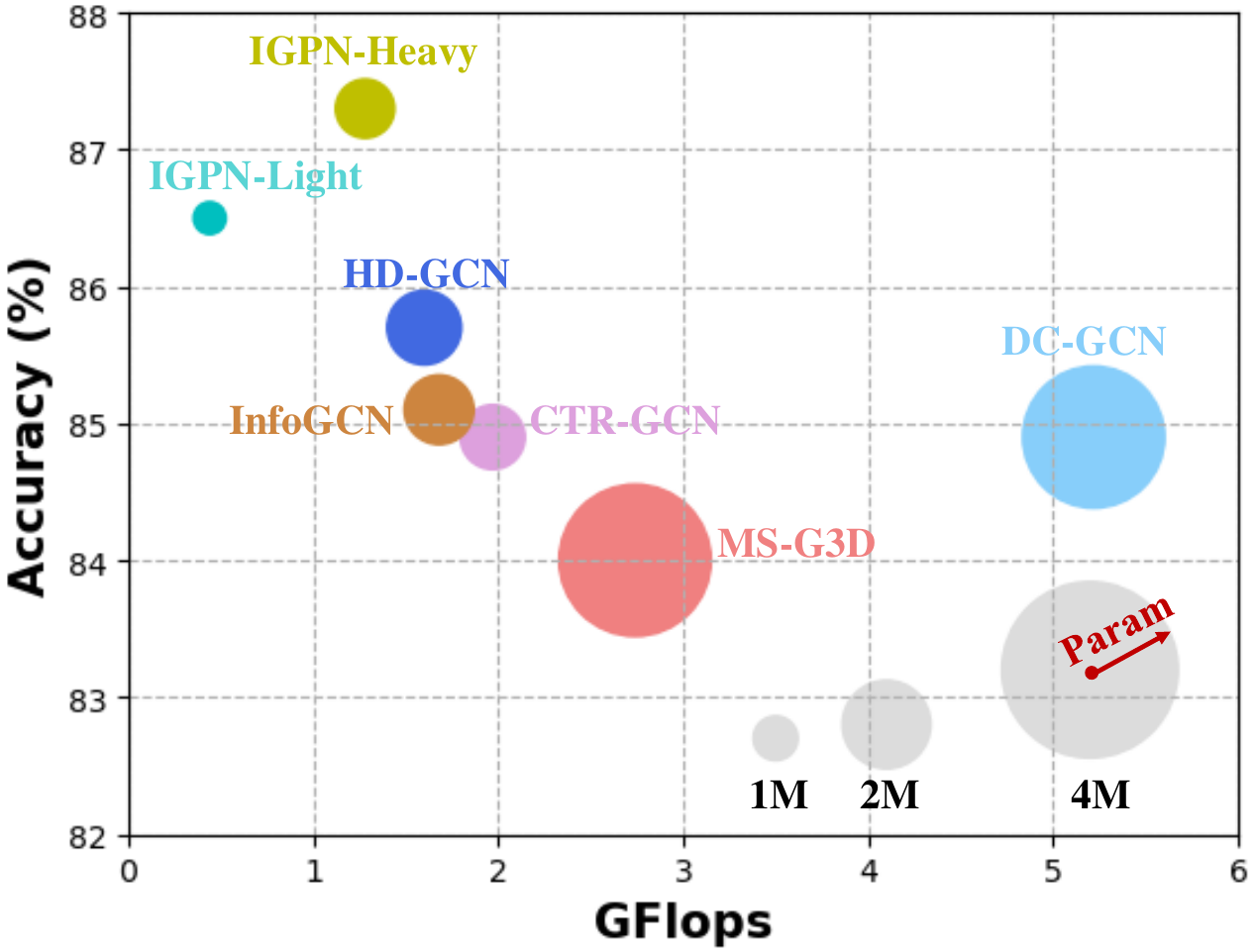}
\caption{
The performance of mainstream algorithms on Cross-Subject of the NTU-RGB+D 60 dataset. 
(All results are obtained from the joint-stream network.)
}
\label{fig4}
\end{figure}

\begin{table}[t]
\caption{Accuracy (\%) comparison of our approach 
under multi-stream evaluation
with the state-of-the-art methods on NTU-RGB+D 60\&120.}
\centering
\scalebox{0.92}{
\begin{tabular}{l|c c|c c}
\hline
\multirow{2}{*}{{Methods}}  &\multicolumn{2}{c|}{{NTU-RGB+D 60}} &\multicolumn{2}{c}{{NTU-RGB+D 120}}  \\ \cline{2-5}
         & {X-Sub}         & {X-View} & {X-Sub} & {X-Set} \\
\hline
Lie Group {\cite{vemulapalli2014human}}                   & 50.1          & 52.8    &- &-   \\
H-RNN {\cite{du2015hierarchical}}                       & 59.1          & 64.0   &- &-  \\
PA-LSTM {\cite{shahroudy2016ntu}}                     & 62.9          & 70.3    & 25.5          & 26.3   \\
SkeMotion{\cite{liu2017global}}                    &  -             & -        & 67.7         & 66.9   \\
STA-LSTM {\cite{song2017end}}                      & 73.4          & 81.2    &- &-    \\
Visualize CNN {\cite{liu2017enhanced}}            & 76.0          & 82.6   &- &-   \\
Multi CNN+RotClips {\cite{ke2018learning}}      &  -             & -            & 62.2          & 61.8\\
TSRJI {\cite{caetano2019skeleton}}           &  -             & -           & 67.9          & 62.8  \\
Hybrid {\cite{dhiman2020view}}           & 79.4   & 84.1 &- &-    \\
BPLHM {\cite{zhang2019graph}}               & 85.4          & 91.1   & -   & -  \\
GeomNet {\cite{nguyen2021geomnet}}           & 93.6 & 96.3 & 86.5  & 87.6  \\
\hline
ST-GCN {\cite{yan2018spatial}}             & 81.5          & 88.3    &- &-   \\
AGCN {\cite{shi2019two}}               & 88.5          & 95.1   & 82.9   & 84.9  \\
SGN		 {\cite{zhang2020semantics}}	& 89.0         & 94.5      & 79.2         & 81.5  \\
AGC-LSTM {\cite{si2019attention}}       & 89.2          & 95.0  &- &-   \\
SGCN {\cite{peng2020learning}}            & 89.4  & 95.7 &- &- \\    
MV-HPGNet {\cite{wang2020learning}}     &  -             & -                 & 83.5   & 85.4   \\
MV-IGNet {\cite{wang2020learning}}       &  -             & -          & 83.9   & 85.6  \\
Shift-GCN {\cite{cheng2020skeleton}}       & 90.7  & 96.5  & 85.9          & 87.6    \\ 
MS-G3D {\cite{liu2020disentangling}}     & 91.5          & 96.2  & 86.9          & 88.4    \\
Dynamic GCN {\cite{ye2020dynamic}}      & 91.5  & 96.0 & 87.3  & 88.6\\ 
GCN-HCRF {\cite{liu2020multi}}	 &  90.0             & 95.5   &  -             & -    \\
CD-JBF-GCN {\cite{tu2022joint}} &  89.0             & 95.7   &  -             & -   \\
Sym-GNN {\cite{li2021symbiotic}}                & 90.1       & 96.4    &- &- \\
CTR-GCN {\cite{chen2021channel}}            & 92.4       & {96.8}  & 88.9  & 90.6  \\
InfoGCN \cite{chi2022infogcn} & 93.0        & \textbf{97.1}  & 89.8  & 91.2  \\
FR-Head {\cite{zhou2023learning}}            & 92.8       & 96.8  & 89.5  & 90.9  \\
HD-GCN {\cite{lee2023hierarchically}}            & \textbf{93.0}       &97.0  & 89.8  & 91.2   \\
Koompan Pooling {\cite{wang2023neural}}            & 92.9       &96.8  & \textbf{90.0} &\textbf{91.3}   \\
FRF-GCN \cite{yun2024behavioral} & 91.3       &96.5  & 87.1 &88.4   \\
\hline
Graph2Net {\cite{wu2021graph2net}} (Backbone) &  90.1             & 96.0   &  86.0             & 87.6           \\
IGPN-Light (Ours)		&92.2	 &96.4 &89.0 &90.2\\
IGPN-Heavy	 (Ours)		& {92.9}        &  96.7   & {89.6}         &  {91.1}  \\
\hline
\end{tabular}}
\label{table7}
\end{table}

\begin{table*}[h]
\caption{Accuracy (\%) comparison of our approach 
under multi-stream evaluation with the state of the art methods on UWA3D Multiview Activity II Dataset.}
\centering
\scalebox{0.92}{
\begin{tabular}{l|c c|c c|c c|c c|c c|c c|c}
\hline
{Training views} &\multicolumn{2}{c|}{$V_{1}\& V_{2}$} &\multicolumn{2}{c|}{$V_{1}\& V_{3}$} &\multicolumn{2}{c|}{$V_{1}\& V_{4}$} &\multicolumn{2}{c|}{$V_{2}\& V_{3}$} &\multicolumn{2}{c|}{$V_{2}\& V_{4}$} &\multicolumn{2}{c|}{$V_{3}\& V_{4}$} &\multirow{2}{*}{{Mean}}\\  \cline{1-13}
{Test view} &$V_{3}$ &$V_{4}$ &$V_{2}$ &$V_{4}$ &$V_{2}$ &$V_{3}$ &$V_{1}$ &$V_{4}$ &$V_{1}$ &$V_{3}$ &$V_{1}$ &$V_{2}$\\ \hline 
HOJ3D{\cite{xia2012view}}  &15.3&28.2&17.3&27.0&14.6&13.4&15.0&12.9&22.1&13.5&20.3&12.7&17.7\\ 
Actionlet ensemble{\cite{wang2013learning}} &45.0&40.4&35.1&36.9&34.7&36.0&49.5&29.3&57.1&35.4&49.0&29.3&39.8\\ 
Lie Group {\cite{vemulapalli2014human}} &49.4&42.8&34.6&39.7&38.1&44.8&53.3&33.5&53.6&41.2&56.7&32.6&43.4\\ 
Enhanced skeleton visualization{\cite{liu2017enhanced}} &66.4&68.1&56.8&66.1&58.8&66.2&74.2&67.0&76.9&64.8&72.2&54.0&66.0\\ 
Ensemble TS-LSTM v2{\cite{lee2017ensemble}} &72.1&79.1&74.0&77.6&75.6&70.1&79.6&79.9&83.9&66.1&79.2&69.7&75.6 \\
ShiftGCN{\cite{cheng2020skeleton}} &73.3&85.8&71.6&82.3&75.6&74.1&86.3&82.3&83.1&74.1&85.1&{74.4}&79.0\\
ShiftGCN++{\cite{cheng2021extremely}} &{74.5}&\textbf{86.6}&76.8&85.0&75.2&73.7&86.3&82.7&85.5&73.3&84.7&74.0&79.9\\
MCMT-Net \cite{wu2023motion} &\textbf{76.1} &86.0 &\textbf{78.0} &84.6 &76.8 &77.3 &86.3 &81.9 &85.1 &\textbf{74.9} &87.1 &\textbf{74.4} &80.7\\
\hline
IGPN-Heavy (Ours) &74.1 &84.3 &{76.8} &\textbf{86.6} &\textbf{78.7} &\textbf{78.9} &\textbf{86.7} &\textbf{84.6} &\textbf{87.1} &{74.5} &\textbf{88.6} &73.6 &\textbf{81.2}\\
\hline
\end{tabular}}
\label{table8}
\end{table*}

\subsection{A Comparison with the state-of-the-art}
The effectiveness of the proposed method is verified through comprehensive comparisons with state-of-the-art methods on multiple benchmarks, as presented in Figure~\ref{fig4}, Table~\ref{table7} and Table~\ref{table8}.

First, we conducted a performance comparison between IGPN and mainstream methods under single-stream evaluation.
As shown in Figure~\ref{fig4}, the proposed IGPN outperforms HD-GCN~\cite{lee2023hierarchically}, InfoGCN~\cite{chi2022infogcn}, CTR-GCN~\cite{chen2021channel}, MS-G3D~\cite{liu2020disentangling}, DC-GCN~\cite{cheng2020decoupling} in accuracy while maintain fewer Flops and parameters.
These results illustrate the efficiency and effectiveness of the proposed method, which destined its usability in real applications.

From Table~\ref{table7}, the results show demonstrate the proposed method is competitive with state-of-the-art methods. We observe that,  compared with the baseline method Graph2Net, both version of IGPN achieve significant improvement in accuracy on NTU-RGB+D 60 and NTU-RGB+D 120 datasets. 
CTR-GCN is one of the most representative methods published in recent years; 
compared with this method, IGPN-Heavy maintains a certain accuracy advantage. 
Following CTR-GCN, recent methods have shown limited improvements in accuracy. Compared with these methods, our method still achieves comparable performance.
On the UWA3D dataset, 
From Table~\ref{table8}, 
IGPN-Heavy outperforms previous methods in 7 out of 12 possible combinations, with a higher average performance when compared with previous state-of-the-art.

Our method has achieved a clear lead in both accuracy and efficiency under single-stream evaluation. Under multi-stream evaluation, our method has also achieved performance similar to or better than the optimal method in terms of accuracy.

\begin{table}[t]
\centering
\caption{The impact of different modelling strategies.}
\scalebox{0.92}{
\begin{tabular}{c c c c c c}
\hline
{Method}	& {Adaptive Pooling} & {CFB}  & {ISM}    & {Acc (\%)}  \\
\hline
{baseline} & \small \XSolidBrush	& \small \XSolidBrush & \small \XSolidBrush 	    & 88.8 \\
\hline
{with SGP~\cite{chen2020graph}} & \small \XSolidBrush	& \small \XSolidBrush & \small \XSolidBrush 	     & 88.3 \\
\hline
{\multirow{4}{*}{with IGPN}} 
     &\small \Checkmark & \small \XSolidBrush & \small \XSolidBrush    & 89.1 \\
          &\small \Checkmark &\small \Checkmark & \small \XSolidBrush  & 89.9 \\
&\small \Checkmark & \small \XSolidBrush &\small \Checkmark     &90.2  \\
&\small \Checkmark &\small \Checkmark &\small \Checkmark   & {91.2} \\
\hline
\end{tabular}}
\label{table2}
\end{table}

\subsection{Ablation Studies}
In this section, we present a comprehensive set of experiments and analyses for all of our proposed innovations on cross-subject of NTU-RGB+D 60.

\noindent\textbf{Effects of different modules on performance.}
To demonstrate the effectiveness of our innovations, we have chosen Graph2Net~\cite{wu2021graph2net} with same number of layers as the baseline for comparison. As shown in Table~\ref{table2}, we conducted a thorough analysis of our proposed modules and found that the adaptive mechanism improved the model's accuracy to 89.1\%. The fusion module further improved the accuracy by 0.8\%. Finally, after supplementing the input features, the recognition performance reached 91.2\%. The experimental results indicate a steady improvement in recognition accuracy by gradually adding our innovations. The final accuracy was significantly better than that of the original model and SGP.

\noindent\textbf{Performance comparison on different backbones.}
In this section, we investigate the applicability of our pooling strategy to other widely used methods, including AGCN~\cite{shi2019two}, Graph2Net~\cite{wu2021graph2net} and CTR-GCN~\cite{chen2021channel}. Specifically, we keep the basic framework unchanged and only replace specific graph convolution operations with these methods separately. The results in Table~\ref{table_} consistently demonstrate that the proposed improved pooling strategy can enhance model accuracy while reducing computational overhead significantly.
We choose the Graph2Net as the default graph convolution operation for all other experiments.

\noindent\textbf{Analysis of specific configurations.}
In this section, we will conduct detailed experiments and analyses on the parameter selection and structural design.
\begin{table}[t]
\caption{Configurations analysis in the pooling strategy.}
\centering
\scalebox{0.92}{
\begin{tabular}{c c c c c}
\hline
{Method}            & {$r$} & {$\sigma$} & {Location} & {Acc (\%)} \\
\hline
\multirow{7}{*}{ASGP} & 2  & Tanh     &$1-3$     & 88.4    \\ 
                      & \cellcolor{lightgray}4  & \cellcolor{lightgray}Tanh     &\cellcolor{lightgray}$1-3$     & \cellcolor{lightgray}{89.1}    \\ 
                      & 8  & Tanh     &$1-3$     & 88.6    \\ 
                      & 4  & Sigmoid  &$1-3$     &  88.3   \\
                      & 4  & SoftMax  &$1-3$     & 88.4    \\ 
                      & 4  & Tanh     &$1$       & 88.5    \\ 
                      & 4  & Tanh     &$1-2$     &  88.6   \\ 
\hline
ASGP $wo.$ Res        & 4  & Tanh   &$1-3$       & 88.6  \\
\hline
\end{tabular}}
\label{table3}
\end{table}
\begin{table}[t]
\caption{Configurations analysis in cross fusion block.}
\centering
\scalebox{0.92}{
\begin{tabular}{c c c c}
\hline
{$s$} &{Location}	& {Fusion} & {Acc (\%)}  \\
\hline
0.25 &$1-3$ & Sum & 89.5 \\
\rowcolor{lightgray}0.5  &$1-3$ & Sum  & {89.9}\\
0.75 &$1-3$ & Sum & 89.6 \\
0.5 &$1$ & Sum & 89.0 \\
0.5 &$1-2$ & Sum & 89.5 \\
0.5  &$1-3$ & Concat  & 89.7\\
\hline
\end{tabular}}
\label{table4}
\end{table}
\begin{table}[t]
\caption{Configurations analysis in information supplement module.}
\centering
\scalebox{0.92}{
\begin{tabular}{c c c c c}
\hline
{BatchNorm}	& {Embed} & {layer} & {channel} & {Acc (\%)}  \\
\hline
\small \XSolidBrush & GraphConv	&2  &32	  & 90.1 \\
\rowcolor{lightgray}\Checkmark  & GraphConv	&2 	&32  & {91.2} \\
\small \Checkmark  & 2D Conv & 2  &32  & 90.7 \\
\small \Checkmark  & GraphConv	&1 	&32  & 90.9 \\
\small \Checkmark  & GraphConv	&3 	&32  & 91.0 \\
\small \Checkmark  & GraphConv	&2 	&16  & 90.9 \\
\small \Checkmark  & GraphConv	&2 	&64  & 90.9 \\
\hline
\end{tabular}}
\label{table5}
\end{table}

\begin{itemize}
\item {Structure pooling strategy with region awareness.} 
As mentioned previously, we utilise SGP with the new implementation as the baseline to verify the impact of the normalisation function $\sigma$, the parameters of projection space $r$, as well as the location of the pooling operation. The experimental results are presented in Table~\ref{table3}.
Firstly, we observe that the recognition accuracy improves after incorporating the adaptive mechanism. Moreover, our model performs best when $r=4$ and $\operatorname{Tanh}$ is used as the normalisation function. 
Secondly, the model achieves an accuracy of 89.1\% when the adaptive mechanism is introduced in all three pooling operations. 
Our also investigate the impact of retaining the initially assigned matrix through residual connection when introducing the adaptive mechanism. Results showed that removing the residual mechanism led to a drop in performance to 88.6\%, highlighting the importance of preserving fundamental structural properties while introducing flexibility.

\item {Cross fusion block.} 
In our experiments, we investigated the impact of feature augmentation on performance when incorporating the new pooling strategy. Firstly, we explored the effect of the weight parameter in Equation~\ref{eqno12}. The results in Table~\ref{table4} show that the best performance is achieved when the weights of the two features being fused are both 0.5.
After fixing the weight $s$, we further investigated the optimal location and fusion strategy for the fusion module. The experimental results show that the performance improves as more fusion modules are added, ultimately reaching 89.9\%. We also found that summation performs better than concatenation in the channel dimension. Notably, we perform the fusion after the global pooling operation for the last fusion.

\item {Information supplement module.} 
The experiments conducted in this section aimed to investigate the impact of parameter selection and structural design on the performance of the model. 
The results from Table~\ref{table5} show that batch normalisation on both position-based and vector-based features is necessary for better performance.
Also, changing the graph-based embedding operation to regular 2D convolution embedding resulted in a 0.5\% decrease in performance. Furthermore, the optimal structure for graph embedding is found to be two layers of graph convolution with a final projected space dimension of 32.
\end{itemize}

\begin{figure}[t]
\centering
\subfloat[]{\includegraphics[width=1in]{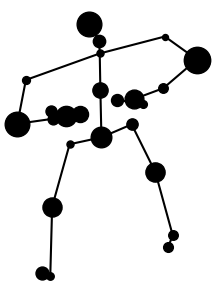}}
\hfil
\subfloat[]{\includegraphics[width=1in]{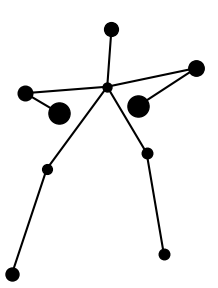}}
\hfil
\subfloat[]{\includegraphics[width=1in]{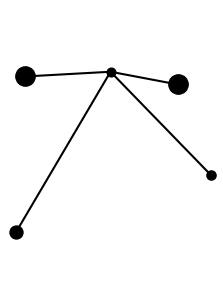}}
\hfil
\subfloat[]{\includegraphics[width=1in]{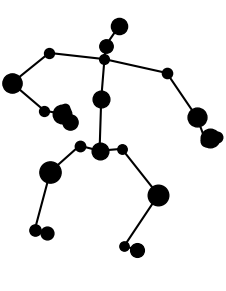}}
\hfil
\subfloat[]{\includegraphics[width=1in]{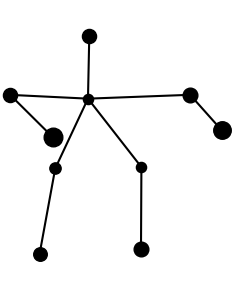}}
\hfil
\subfloat[]{\includegraphics[width=1in]{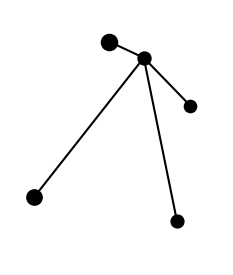}}
\hfil\caption{\textbf{Visualisation of region awareness.} We select two typical behaviours and visualise the associated responses of their nodes. In order to facilitate the display, we perform certain transformations on the specific values corresponding to the node size; that is, the larger the node, the more critical the node is.
(a)(b)(c) represents 'rub two hands', (d)(e)(f) represents 'jump up'.
}
\label{fig5}
\end{figure}

\begin{figure*}[h]
\centering
\includegraphics[width=1.0\textwidth]{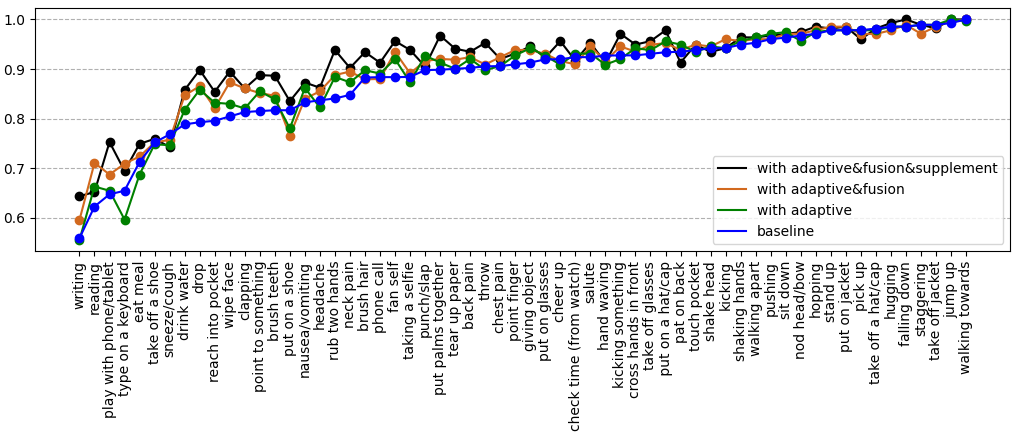}
\caption{
Comparison results for each category on the NTU-RGB+D Cross-Subject dataset.
}
\label{fig:cate}
\end{figure*}

\noindent\textbf{Effects of different modules on performance.}
We have presented the recognition accuracy of the models with the three proposed modules in each category, as shown in Figure~\ref{fig:cate}. Based on the observations and analysis of the figure, several conclusions can be drawn:
The incremental incorporation of innovative modules has a discernible impact on the recognition accuracy of different categories.
For instance, the addition of the adaptive mechanism markedly enhances the recognition accuracy for behaviours such as drop and rub two hands, which necessitate more flexible pooling operations compared to the baseline. On the other hand, the fusion module performs exceptionally well in reading, wipe face, and other behaviours by providing more distinguishable multi-granularity feature representations.
Moreover, the results reveal that the introduction of these two innovations had a certain degree of adverse effects on some categories, such as putting on a shoe and taking a selfie. Fortunately, the supplementary modules has successfully mitigated these adverse effects.

\noindent\textbf{Visualisation of region awareness.}
To provide a more tangible illustration of our novel pooling method, we generated a visualisation of the region perception in  Figure~\ref{fig5}, where the size of each point represents the importance of its corresponding position, reflecting the region that needs to be preserved during the pooling operation.
From (a), we can observe that the response at the hand joints is relatively large for the action 'rub two hands', and this trend is also evident in (b) and (c). For the action 'jump up' in (d) (e) (f), the response of the hands and legs is more evident due to the need for the cooperation of the entire body.
The observed phenomena demonstrate that our model can adaptively enhance information acquisition in critical regions for actions that focus on local areas or require whole-body coordination, providing discriminative feature expression during pooling.

\begin{table}[t]
\caption{The experimental results obtained by the multi-stream fusion.}
\centering
\scalebox{0.92}{
\begin{tabular}{c c c c}
\hline
{Methods}	& {Spatial} & {Motion}  & {Acc (\%)}  \\
\hline
IGPN-Light & \small \Checkmark	& \small \XSolidBrush 	  & 90.2 \\
IGPN-Light &\small \Checkmark	&\small\Checkmark   &91.6  \\ 
IGPN-Light$_{En}$ &\small\Checkmark &\small \Checkmark  & 92.2 \\  
\hline
IGPN-Heavy &\small \Checkmark	& \small \XSolidBrush 	  & 91.2 \\
IGPN-Heavy &\small \Checkmark	&\small \Checkmark   & 92.3 \\ 
IGPN-Heavy$_{En}$ &\small \Checkmark & \small \Checkmark  & 92.9 \\  
\hline
\end{tabular}}
\label{table6}
\end{table}

\noindent\textbf{Multi-stream fusion.} 
The effectiveness of multi-stream network fusion has been extensively demonstrated in previous research~\cite{shi2019two,cheng2020skeleton,liu2020disentangling,chen2021channel,wu2021graph2net}. We also adopt this approach to improve the classification performance of our algorithm further for a fairer comparison.
First, we separately verify the impact of spatial features and motion features on performance. The results in Table~\ref{table6} show that the recognition effect can be improved to 
91.6\% and
92.3\% by adopting the two-stream fusion strategy.
We also explored the use of an ensemble approach where we trained another separate network using only half of the frames. This technique is commonly used in other works such as~\cite{zolfaghari2018eco,lin2019tsm}. We denoted this model as IGPN$_{En}$. The results in Table~\ref{table6} show that this ensemble mechanism improves the final accuracy to 
92.2\% and
92.9\% respectively.

\begin{table}[t]
\caption{The experimental results obtained by different multi-stream fusion strategies.}
\centering
\scalebox{0.92}{
\begin{tabular}{c c c c}
\hline
{Method}	&Stream &Flops (G) & Acc (\%) \\
\hline
IGPN-Light (no ISM) &J\&B\&JM\&BM &2.05&92.0\\
IGPN-Light  &J\&J$_{En}$\&JM\&JM$_{En}$ &1.33 &92.2 \\
IGPN-Heavy (no ISM)  &J\&B\&JM\&BM &5.38 &92.5\\
IGPN-Heavy &J\&J$_{En}$\&JM\&JM$_{En}$ &3.83 & 92.9 \\
\hline
\end{tabular}}
\label{table:diff}
\end{table}

\noindent\textbf{Effects of different multi-stream strategies} 
Since the designed ISM module includes joint-based and bone-based features, there may be confusion about the difference between this method and training joint-based and bone-based networks separately.
Indeed, mainstream methods\cite{shi2019two,cheng2020skeleton,liu2020disentangling,chen2021channel} generally choose to train separate joint-based, bone-based, and corresponding motion-based features, and fuse the corresponding classification features to obtain the final recognition result of the multi-stream recognition network.
As shown in Table~\ref{table:diff}, we choose to delete the ISM module from the IGPN and train on different feature streams respectively, and finally fuse their results.
The results show that compared with our default method, this strategy not only brings a significant increase in computational effort (2.05GFlops vs 1.33GFlops on IGPN-Light,
5.38GFlops vs 3.83GFlops on IGPN-Heavy), and also leads to a decrease in accuracy performance
(92.0\% vs 92.2\% on IGPN-Light
92.5\% vs 92.9\% on IGPN-Heavy)
.

\section{Conclusion}
Existing skeleton-based graph pooling research suffers from the failure to consider graph properties of the skeleton, resulting in inflexibility and loss of vital information.
Here we have adopted several effective improvement strategies to enhance and supplement the pooling strategy from multiple perspectives.
(1) We comprehensively analysed the strengths and limitations of various traditional graph pooling strategies and devised a pooling strategy that significantly enhances flexibility while preserving the structural characteristics based on the local-global association.
(2) To address the issues of information loss and network under-saturation caused by pooling operations, we propose block-level multi-granularity representations and input-level discriminative information embedding.
On the above innovations, we constructed IGPN. The lightweight version can greatly improve efficiency and effect at the same time; the heavyweight version can further improve classification accuracy by sacrificing certain computing efficiency.

\bibliographystyle{ACM-Reference-Format}
\bibliography{sample-base}

\end{document}